\documentclass[conference, fleqn]{IEEEtran}
\IEEEoverridecommandlockouts
\usepackage{cite}
\usepackage{algorithm}
\usepackage{amsmath,amssymb,amsfonts}
\usepackage{algorithmic}
\usepackage{wrapfig}
\usepackage{graphicx}
\usepackage{textcomp}
\usepackage{changepage}
\usepackage{algorithm}
\usepackage{listings}
\usepackage{colortbl}
\usepackage{bbding}
\usepackage{booktabs}
\usepackage[table,xcdraw]{xcolor}
\usepackage{multirow}
\usepackage{orcidlink} 
\usepackage{colortbl}
\usepackage{hyperref}
\def\BibTeX{{\rm B\kern-.05em{\sc i\kern-.025em b}\kern-.08em
    T\kern-.1667em\lower.7ex\hbox{E}\kern-.125emX}}
\usepackage[normalem]{ulem}
\useunder{\uline}{\ul}{}

\begin{document}

\title{HGMamba: Enhancing 3D Human Pose Estimation with a HyperGCN-Mamba Network}
\author{\IEEEauthorblockN{Anonymous Authors}}
\author{\IEEEauthorblockN{Hu Cui\,\orcidlink{0000-0002-6431-5939}}
\IEEEauthorblockA{\textit{Information and Management Systems Engineering} \\
\textit{Nagaoka University of Technology}\\
Nagaoka-shi, Japan \\
s227006@stn.nagaokaut.ac.jp}
\and
\IEEEauthorblockN{Tessai Hayama*\thanks{* Corresponding author} \thanks{This work was supported by JSPS KAKENHI Grant Number
23K11364.}}
\IEEEauthorblockA{\textit{Information and Management Systems Engineering} \\
\textit{Nagaoka University of Technology}\\
Nagaoka-shi, Japan \\
t-hayama@kjs.nagaokaut.ac.jp}
}


\maketitle

\begin{abstract} 3D human pose lifting is a promising research area that leverages estimated and ground-truth 2D human pose data for training. While existing approaches primarily aim to enhance the performance of estimated 2D poses, they often struggle when applied to ground-truth 2D pose data.
We observe that achieving accurate 3D pose reconstruction from ground-truth 2D poses requires precise modeling of local pose structures, alongside the ability to extract robust global spatio-temporal features. To address these challenges, we propose a novel Hyper-GCN and Shuffle Mamba (HGMamba) block, which processes input data through two parallel streams: Hyper-GCN and Shuffle-Mamba.
The Hyper-GCN stream models the human body structure as hypergraphs with varying levels of granularity to effectively capture local joint dependencies. Meanwhile, the Shuffle Mamba stream leverages a state space model to perform spatio-temporal scanning across all joints, enabling the establishment of global dependencies. By adaptively fusing these two representations, HGMamba achieves strong global feature modeling while excelling at local structure modeling.
We stack multiple HGMamba blocks to create three variants of our model, allowing users to select the most suitable configuration based on the desired speed-accuracy trade-off. Extensive evaluations on the Human3.6M and MPI-INF-3DHP benchmark datasets demonstrate the effectiveness of our approach. HGMamba-B achieves state-of-the-art results, with P1 errors of 38.65 mm and 14.33 mm on the respective datasets. Code and models are available: \url{https://github.com/HuCui2022/HGMamba}
\end{abstract}


\section{Introduction}
3D Human Pose Estimation (HPE) in videos focuses on predicting the positions of human body joints in 3D space. This task has numerous applications, including video surveillance \cite{baltieri20113dpes, cui2024stsd, chung2022comparative}, human-computer interaction \cite{chen2021channel, cui2022lgst, yan2018spatial, cui2024joint}, and physical therapy \cite{sarafianos20163d}. As the range of these applications continues to grow, there is an increasing demand for pose estimation models that offer both high accuracy and computational efficiency.
In most real-world scenarios, pose sequences are captured in 2D, largely due to the widespread use of standard RGB cameras. Consequently, a critical challenge in this domain is effectively and efficiently lifting these 2D pose sequences into 3D space while maintaining precision and robustness.


The advantages of the 2D-to-3D lifting approach can be summarized in two key aspects: leveraging advancements in 2D human pose detection and utilizing temporal information across multiple 2D pose frames. 
Significant progress in 2D-to-3D lifting for estimated 2D human pose data has been achieved through advanced detectors such as Mask R-CNN (MRCNN) \cite{he2017mask}, Cascaded Pyramid Networks (CPN) \cite{chen2018cascaded}, Stacked Hourglass (SH) \cite{stackedhourglass}, HRNet \cite{sun2019deep}, YOLO-Pose \cite{maji2022yolo}, and Mediapipe \cite{lugaresi2019mediapipe}. The intermediate stage of 2D pose estimation using these detectors substantially reduces the data volume and complexity of the 3D HPE task. Furthermore, better 3D HPE results can be achieved by employing high-accuracy 2D detectors \cite{zhang2022mixste} or fine-tuning 2D pose estimations to align more closely with ground truth data \cite{yu2023gla}.
Temporal information also plays a crucial role in 2D-to-3D lifting. State-of-the-art methods have demonstrated significant improvements by leveraging long sequences of 2D poses. Recent works \cite{li2022mhformer, einfalt2023uplift, tang20233d, zhao2023poseformerv2, qian2023hstformer, chen2023hdformer, zhu2023motionbert, li2022exploiting, mehraban2024motionagformer} harness the power of transformers to capture global temporal dependencies across long 2D pose sequences, leading to remarkable advancements in 3D HPE performance.


For 2D-to-3D lifting methods, recent advanced models can be broadly categorized into three main types: Temporal Convolutional Network (TCN)-based \cite{pavllo20193d, liu2020attention}, Graph Convolutional Network (GCN)-based \cite{hu2021conditional, yu2023gla}, and Transformer-based approaches \cite{li2022mhformer, einfalt2023uplift, tang20233d, zhao2023poseformerv2, qian2023hstformer, chen2023hdformer, zhu2023motionbert, li2022exploiting, mehraban2024motionagformer}.
TCN- and Transformer-based methods excel at capturing long-range dependencies due to their use of strided convolutions and global attention mechanisms, respectively. However, these methods often flatten the 2D pose sequences before feeding them into the model, making it challenging to incorporate intuitive designs that explicitly capture local joint features based on the underlying pose structure.
In contrast, GCN-based models inherently preserve the structural relationships of 2D and 3D human poses during feature propagation. While this structural advantage allows for better representation of joint interactions, existing GCN-based approaches have yet to fully exploit this potential, as they primarily focus on joint-to-joint interactions and neglect the broader physiological structure of human motion.
\begin{figure*}[t]
	\centering
		\centering
		\includegraphics[width=\linewidth]{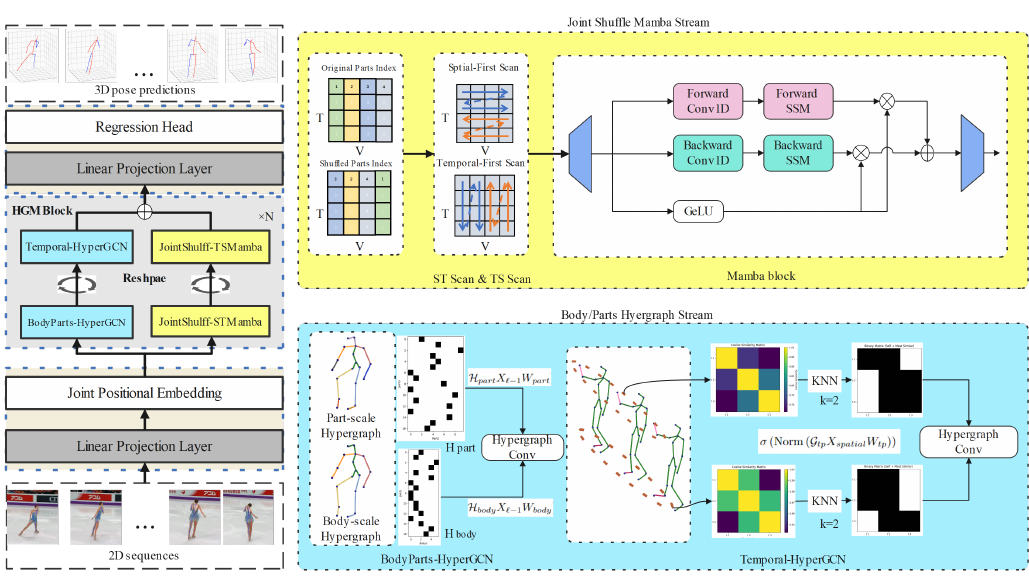}
	\caption{Overview of HGMamba. It is composed of $N$ dual-stream HGM blocks. The Hyper-GCN stream captures local human body part semantics, while the state-space module (Mamba-block) effectively models robust global information.}
	\label{fig:2}
\end{figure*}


To overcome the limitations of existing approaches, we propose a hybrid spatiotemporal architecture, HGMamba, which combines Hyper-GCN \cite{bai2021hypergraph, zhu2022selective} and Mamba Networks \cite{mamba, mamba2, zhu2024vision, pavlakos2018ordinal} for 3D human pose estimation. HGMamba consists of two distinct streams: the Hyper-GCN stream, which focuses on modeling local spatiotemporal information, and the Mamba stream, which captures global spatiotemporal dependencies.


For ground truth 2D pose data or highly accurate 2D estimated pose data, extracting local joint structure information becomes even more crucial due to the absence of 2D noise. To preserve more physiological structure in the local joint information, our Hyper-GCN stream is divided into two components: Part-scale Hyper-GCN and Body-scale Hyper-GCN. In the Body-scale Hyper-GCN, there are five hyperedges representing the torso and four limbs. The Part-scale Hyper-GCN, on the other hand, includes 10 hyperedges, which separate the head, hands, and feet from the torso, arms, and legs. The dynamic fusion of these two local features enhances the model's ability to capture and retain detailed structural information.

For global modeling of 2D or 3D pose sequences, current state-of-the-art approaches \cite{li2022mhformer, einfalt2023uplift, tang20233d, zhao2023poseformerv2, qian2023hstformer, chen2023hdformer, zhu2023motionbert, li2022exploiting, mehraban2024motionagformer} primarily rely on transformer-based methods that use self-attention mechanisms for global spatio-temporal modeling. However, these methods suffer from quadratic complexity when computing the affinity matrix, leading to a significant increase in computational cost as the number of joints and frames grows. In contrast, state-space models (SSMs) \cite{mondal2024hummuss, gu2022parameterization, gu2021efficiently, gupta2022diagonal, mehta2022long} effectively overcome these limitations. Our Mamba stream leverages recent advances in SSMs \cite{mamba2, zhu2024vision}, scanning joints across both intra- and inter-frames and dynamically fusing the two types of global information. To improve its resistance to overfitting, we also incorporate a spatial shuffle mechanism \cite{zhang2018shufflenet, huang2021shuffle, huang2024stochastic}.

We employ adaptive fusion to combine features from both the Hyper-GCN and Mamba streams. This approach ensures that the model captures both local and global representations of human motion, thereby enhancing its performance not only on 2D estimated poses but also on ground-truth 2D poses. 

%


Our main contributions are as follows:

\begin{itemize} \item We introduce HGMamba, a spatio-temporal architecture that combines Hyper-GCN and state-space models (SSMs) for human motion processing, without relying on attention mechanisms. \item We demonstrate the importance of incorporating a fine-grained human structure prior for 3D pose lifting using either 2D ground truth or high-quality estimated 2D pose data. \item HGMamba achieves state-of-the-art performance on two challenging benchmarks, Human3.6M and MPI-INF-3DHP, for 2D ground truth data, while also achieving competitive results with estimated 2D pose data. \end{itemize}


\section{Preliminary}
\noindent\textbf{Selective Structured State Space Models.}  
The State Space Model (SSM) in S4 \cite{gu2021efficiently} is based on the classical state space model, which maps a 1D input \( x(t) \in \mathbb{R} \) to a 1D output \( y(t) \in \mathbb{R} \) through an \( N \)-dimensional latent state \( h(t) \in \mathbb{R}^N \). This process is described by the following system of linear ordinary differential equations (ODEs):
\begin{equation}
     \begin{aligned}
    h^{\prime}(t) &= \mathbf{A} h(t) + \mathbf{B} x(t), \\
    y(t) &= \mathbf{C} h(t) \label{eq-1}
    \end{aligned}
\end{equation}
where \( \mathbf{A} \in \mathbb{R}^{N \times N} \) is the evolution matrix, and \( \mathbf{B} \in \mathbb{R}^{N \times 1} \) and \( \mathbf{C} \in \mathbb{R}^{1 \times N} \) are the projection parameters of the neural network. The term \( h^{\prime}(t) \) represents the derivative of \( h(t) \) with respect to time \( t \).

To handle the discrete input sequence \( x = (x_0, x_1, \dots) \in \mathbb{R}^{L} \), various discretization methods \cite{gu2021efficiently, mamba, mamba2} can be applied to discretize the parameters in Eq. \ref{eq-1} using a step size \( \Delta \) \cite{mamba}. The discretized parameters \( \overline{\mathbf{A}} \) and \( \overline{\mathbf{B}} \) are given by:
\begin{equation}
    \begin{aligned}
    \overline{\mathbf{A}} &= \exp(\Delta \mathbf{A}), \\
    \overline{\mathbf{B}} &= (\Delta \mathbf{A})^{-1} \left( \exp(\Delta \mathbf{A}) - \mathbf{I} \right) \cdot \Delta \mathbf{B}.
    \end{aligned} \label{eq-2}
\end{equation}
After discretizing \( \mathbf{A} \) and \( \mathbf{B} \) to \( \overline{\mathbf{A}} \) and \( \overline{\mathbf{B}} \), Eq. \ref{eq-1} can be reformulated as:
\begin{equation}
    \begin{aligned}
    h_t &= \overline{\mathbf{A}} h_{t-1} + \overline{\mathbf{B}} x_t, \\
    y_t &= \mathbf{C} h_t.
    \end{aligned} \label{eq-3}
\end{equation}
Eq. \ref{eq-3} represents a sequence-to-sequence mapping from \( x_t \) to \( y_t \), which is computed as an RNN. However, due to its sequential nature, this discretized SSM is inefficient for training. To enable efficient parallelizable training, the recursive process can be reformulated as a convolution \cite{gu2021efficiently}:
\begin{equation}
    \begin{aligned}
    \overline{\mathbf{K}} &= \left( \mathbf{C} \overline{\mathbf{B}}, \mathbf{C} \overline{\mathbf{A B}}, \dots, \mathbf{C} \overline{\mathbf{A}}^{L-1} \overline{\mathbf{B}} \right), \\
    \boldsymbol{y} &= \boldsymbol{x} * \overline{\mathbf{K}},
    \end{aligned} \label{eq-4}
\end{equation}
where \( L \) denotes the length of the input sequence \( \boldsymbol{x} \), and \( * \) represents the convolution operation. The convolution kernel \( \overline{\mathbf{K}} \in \mathbb{R}^{L} \) can be efficiently computed using fast Fourier transforms (FFTs).

The parameters in Eq. \ref{eq-1}, Eq. \ref{eq-3}, and Eq. \ref{eq-4} are invariant with respect to the input and temporal dynamics. This linear time-invariant (LTI) property is a fundamental limitation of State Space Models (SSMs) when applied to context-based reasoning \cite{mamba}.
To address this challenge, Mamba \cite{mamba} introduces a selection mechanism by making the parameters \( \mathbf{B}, \mathbf{C}, \Delta \) of the SSM functions of the input sequence \( \boldsymbol{x} \). This design enables effective input-dependent interactions across the sequence.
The selection mechanism in Mamba defines the parameters as follows:
\begin{equation}
    \begin{aligned}
    \mathbf{B} &= \text{LinearN}(x), \\
    \mathbf{C} &= \text{LinearN}(x),
    \end{aligned}
\end{equation}
where both \( \mathbf{B} \) and \( \mathbf{C} \) project the input \( x \) into a dimension \( N \).
Additionally, the parameter \( \Delta \) is defined as:
\begin{equation}
    \begin{aligned}
    \Delta &= \tau_{\Delta}\left(\text{Parameter} + s_{\Delta}(\boldsymbol{x}) \right), \\
    s_{\Delta}(\boldsymbol{x}) &= \text{Broadcast}_D\left(\text{Linear}_1(\boldsymbol{x})\right),
    \end{aligned}
\end{equation}
where the input is first projected to dimension 1 and then broadcast to dimension \( D \). The softplus function is represented as \( \tau_\Delta \).
When the input \( x \) is shaped as \( x \in \mathbb{R}^{B \times L \times D} \), where \( B \) denotes the batch size, \( L \) is the sequence length, and \( D \) is the number of channels, the parameters \( \mathbf{B} \) and \( \mathbf{C} \) are shaped as \( \mathbb{R}^{B \times L \times N} \), and \( \Delta \) is shaped as \( \mathbb{R}^{B \times L \times D} \).

\noindent\textbf{Graph and Hypergraph Convolution.}
The commonly used graph convolutional operation is defined as follows:
\begin{equation}
\begin{aligned}
    \mathrm{GCN}(x) &= \sigma\left(\mathrm{Norm}\left(\tilde{D}^{-\frac{1}{2}}\tilde{A}\tilde{D}^{-\frac{1}{2}}xW_{1}\right)\right), \\
    &= \sigma\left(\mathrm{Norm}\left(\mathcal{G}xW_{1}\right)\right)
\end{aligned} \label{eq-gcn}
\end{equation}
Here, \( \tilde{A} = A + I \) represents the adjacency matrix augmented with self-connections, where \( \tilde{D}_{ii} = \sum_j \tilde{A}_{ij} \) is the degree matrix, obtained by summing the elements of \( \tilde{A} \) along its diagonal. \( W_1 \) is the learnable weight matrix, while \( \text{Norm}(\cdot) \) and \( \sigma(\cdot) \) denote Batch Normalization and the ReLU activation function, respectively. \( \mathcal{G} \) represents the graph convolution kernel.

Based on the original GCN, when graph convolution is extended to hypergraphs \cite{feng2019hypergraph, jiang2019dynamic, bai2021hypergraph, zhu2022selective}, the hypergraph convolution operation can aggregate skeleton information across multiple scales. The hypergraph convolution is defined as follows:
\begin{equation}
\begin{aligned}
    \mathrm{HGCN}(x) &= \sigma\left(\mathrm{Norm}\left(D_v^{-\frac{1}{2}} HM D_e^{-1} H^{\top} D_v^{-\frac{1}{2}} x W_2 \right)\right), \\
    &= \sigma\left(\mathrm{Norm}\left(\mathcal{H} x W_2\right)\right),
\end{aligned} \label{eq-hgcn}
\end{equation}
where \( H \) denotes the hypergraph incidence matrix, \( D_e \) is the diagonal matrix of hyperedge degrees, and \( D_v \) is the diagonal matrix of vertex degrees. \( M \) is a learnable diagonal matrix of hyperedge weights, initially set to the identity matrix (i.e., equal weights for all hyperedges). \( \mathcal{H} \) represents the kernel of Hyper-GCN.


\section{HGMamba}
\subsection{Overall Architecture}
Fig. \ref{fig:2} provides an overview of the proposed HGmamba, which consists of three main components: joint positional embedding, stacked HGM blocks, and the regression head. The positional embedding maps the input 2D coordinates of each joint into a high-dimensional feature space. The HGM blocks aggregate both local and global spatio-temporal contexts, refining the representation of each joint. Using the learned features, the regression head estimates the 3D coordinates.

\noindent \textbf{Joint Positional Embedding}.
Given a 2D pose sequence \( P_{2D} \in \mathbb{R}^{T \times J \times 2} \), where \( T \) denotes the number of frames and \( J \) represents the number of body joints in each frame, we first project \( P_{2D} \) to a high-dimensional space using a linear projection layer. Then, we add the positional encoding \( P_{pos} \in \mathbb{R}^{T \times J \times D} \) to the projected feature \( X_0 \in \mathbb{R}^{T \times J \times D} \), which serves as the input to the first HGM block.

\noindent \textbf{HGM Blocks.}
The HGM framework effectively captures the underlying 3D structure of the skeletal sequence through a dual-stream architecture consisting of the Mamba stream and the HyperGCN stream. The Mamba stream models global dependencies by leveraging its efficient long-range modeling capability, while the HyperGCN stream captures local semantic information by incorporating the human body structure prior. This dual-stream design ensures comprehensive and accurate 3D pose estimation. Further details are provided in Section \ref{sec-block}.

\noindent \textbf{Regression Head.}
In the regression head, we first project the output feature of the $N$-th HGM block, denoted as $X_N \in \mathbb{R}^{T \times J \times D}$, to a higher-dimensional space through a linear layer. A linear regression head is then used to estimate the 3D pose coordinates, $\hat{P}_{3D} \in \mathbb{R}^{T \times J \times 3}$. 

Similar to previous works \cite{chen2023hdformer, zhu2023motionbert, mehraban2024motionagformer}, the entire architecture is optimized by minimizing both the position loss $\mathcal{L}_{3D}$ and the velocity loss $\mathcal{L}_{v}$:
\begin{equation}
    \mathcal{L} = \mathcal{L}_{3D} + \lambda \mathcal{L}_{v},
\end{equation}
\begin{equation}
    \begin{aligned}
        &\mathcal{L}_{3D} = \sum_{t=1}^{T} \sum_{j=1}^{J} \left\| P_{3D}(t,j) - \hat{P}_{3D}(t,j) \right\|, \\
        &\mathcal{L}_{v} = \sum_{t=2}^{T} \sum_{j=1}^{J} \left\| \Delta_{P_{3D}(t,j)} - \Delta_{\hat{P}_{3D}(t,j)} \right\|, 
    \end{aligned}
\end{equation}
where $\lambda$ is a weighting factor. Here, $\Delta_{P_{3D}(t,j)} = P_{3D}(t,j) - P_{3D}(t-1,j)$ and $\Delta_{\hat{P}_{3D}(t,j)} = \hat{P}_{3D}(t,j) - \hat{P}_{3D}(t-1,j)$, with $P_{3D}$ representing the ground truth 3D pose coordinates.

\subsection{HGMamba Block} \label{sec-block}
As shown in Fig. \ref{fig:2}, to obtain expressive spatio-temporal structural features, both Mamba and HyperGCN streams need to model the skeleton sequences from spatial and temporal perspectives, respectively.

\noindent \textbf{Shuffled Mamba Stream.}
The spatial and temporal components of the Mamba stream share the same block structure, as shown in the yellow block of Fig. \ref{fig:2}. Formally, given a batched input $X_{\ell-1} \in \mathbb{R}^{B \times T \times J \times D}$ for the $\ell$-th HGM block, the spatial-first selective Mamba reshapes it to $X_{\ell-1} \in \mathbb{R}^{B \times (L = TJ) \times D}$, while the temporal-first selective Mamba reshapes it to $X_{\ell-1} \in \mathbb{R}^{B \times (L = JT) \times D}$. The Mamba block starts with layer normalization and consists of three distinct pathways. The first pathway processes information independently, while the other two combine it forward and backward across the sequence dimension:
\begin{equation}
    \begin{aligned}
        & X_{id} = \sigma(\text{LayerNorm}(X_{\ell-1} W_{id})), \\
        & X_{f} = SSM_{f}(\sigma(X_{\ell-1} W_{f}^1) W_f^{2}), \\
        & X_{b} = \text{flip}(SSM_{b}(\sigma(\text{flip}(X_{\ell-1}) W_{b}^1) W_b^{2})),
    \end{aligned}
\end{equation}
where $\sigma(\cdot)$ denotes the GeLU activation function \cite{hendrycks2016gaussian}, and $\text{flip}(\cdot)$ denotes flipping along the $L$ dimension. The weight matrices $W_{id}, W_{f}^1, W_{b}^1 \in \mathbb{R}^{D \times nD}$ and $W_f^{2}, W_b^{2} \in \mathbb{R}^{nD \times nD}$ are learnable parameters. We then combine the forward and backward aggregated information using multiplicative gating and reduce the dimension:
\begin{equation}
    \begin{aligned}
        & X^{m}_{\ell-1} = X_{\ell-1} + (X_{id} \odot X_{f} + X_{id} \odot X_{b}) W_{out},
    \end{aligned}
\end{equation}
where $W_{out} \in \mathbb{R}^{nD \times D}$ is the output projection matrix, and $\odot$ denotes the Hadamard product. The Mamba stream can then be expressed as:
\begin{equation}
\begin{aligned}
    & X^m_{\ell-1} = S^{\ell}_m(X_{\ell-1}).
\end{aligned}
\end{equation}
Additionally, we observe that scanning the joints in a fixed spatial order may introduce causal bias, as the joints in a single frame do not have a causal structure. Motivated by previous work \cite{zhang2018shufflenet, huang2021shuffle, huang2024stochastic}, we propose a layer-wise shuffle strategy for training. We use a Bernoulli random variable $b_\ell \in \{0, 1\}$ to indicate whether the $\ell$-th layer should be trained with shuffle regularization. 

The probability $p_\ell$ of the $\ell$-th layer being shuffled is given by:
\begin{equation}
    \begin{aligned}
        & P(b_\ell) = \frac{\ell}{N} P_N,
    \end{aligned}
\end{equation}
where $P_N$ is a constant factor. Algorithm \ref{al-1} provides the details of the shuffled spatial-temporal Mamba implementation.

\begin{algorithm}[H]
\caption{Shuffled Spatial-temporal Mamba}\label{al-1}
\small
\begin{algorithmic}[1]
\label{alg:algo1}
\REQUIRE{feature sequence $\boldsymbol{X}_{\ell-1}\in \mathbb{R}^{B\times T \times J \times D}$, \\layer $S^{\ell}_m$, probability $p_\ell$, training flag $F$, scan mod.}
\ENSURE{token sequence $\boldsymbol{X}_{\ell-1}^{m}$}
\STATE \textcolor{gray}{\text{\# this layer is trained with regularization}}
\IF{$F$ and rand(1) $<$ $p_\ell$ }
    \STATE shuffle\_indices = randperm(J).expand(B, T, J D)
    \STATE restore\_indices = argsort(shuffle\_indices, dim=2)
    \STATE $\boldsymbol{X}_{\ell-1}^{'}$ = gather($\boldsymbol{X}_{\ell-1}$, 2, shuffle\_indices)
\IF{scan mod == 'st'}
    \STATE $\boldsymbol{X}_{\ell-1}^{'} = rearrange(\boldsymbol{X}^{'}_{\ell-1}, b t j d \rightarrow b (t j) d)$
    \STATE $\boldsymbol{X}_{\ell-1}^{'}$ = $S^{\ell}_m$($\boldsymbol{X}_{\ell-1}^{'}$)
    \STATE $\boldsymbol{X}_{\ell-1}^{'} = rearrange(\boldsymbol{X}^{'}_{\ell-1}, b (t j) d \rightarrow b t j d)$
\ELSIF{scan mod == 'ts'}
    \STATE $\boldsymbol{X}_{\ell-1}^{'} = rearrange(\boldsymbol{X}^{'}_{\ell-1}, b t j d \rightarrow b (j t) d)$
    \STATE $\boldsymbol{X}_{\ell-1}^{'}$ = $S^{\ell}_m$($\boldsymbol{X}_{\ell-1}^{'}$)
    \STATE $\boldsymbol{X}_{\ell-1}^{'} = rearrange(\boldsymbol{X}^{'}_{\ell-1}, b (j t) d \rightarrow b t j d)$
\ENDIF
\STATE $\boldsymbol{X}_{\ell-1}^{m}$ = gather($\boldsymbol{X}_{\ell-1}^{'}$, 2, restore\_indices)
\ELSE
\STATE \textcolor{gray}{\text{\# inference or trained without regularization}}
\STATE $\boldsymbol{X}_{\ell-1}^{m}$ = $S^{\ell}_m$($\boldsymbol{X}_{\ell-1}$)
\ENDIF
\STATE 
Return: $\boldsymbol{X}_{\ell-1}^{m}$
\end{algorithmic}
\end{algorithm}


\noindent \textbf{HyperGCN Stream.}
In contrast to Mamba streams, the spatial and temporal feature aggregation in the HyperGCN stream is performed differently. During the local spatial feature modeling process, as shown in the blue block of Fig. \ref{fig:2}, we construct two types of hypergraphs: the body-scale hypergraph and the parts-scale hypergraph, which are based on the dynamic chain structure of the human body.

\begin{equation}
    \begin{aligned}
b1 = &p1,p2 =\{hip, spine, throax,neck\} \{head\} \\
b2 = &p3,p4 =\{right\_hip, right\_knee\} \{right\_feet\} \\
b3 = &p5,p6 =\{left\_hip, left\_knee\} \{left\_feet\}\\
b4 = &p7,p8=\{right\_shoulder, right\_elbow\} \{ right\_wrist\}\\
b5 = &p9,p0= \{left\_sholder, left\_elbow\} \{left\_wrist\}\\
\end{aligned}
\end{equation}
where $b_i$ denotes the joints contained in the body-scale hyperedge of the hypergraph $H_{body}$, and $p_i$ denotes the joints in the part-scale hyperedge of the hypergraph $H_{part}$.

Given a batched input $X_{\ell-1} \in \mathbb{R}^{B \times T \times J \times D}$ for the $\ell$-th spatial HyperGCN block, we first reshape it to $X_{\ell-1} \in \mathbb{R}^{B \times (DT) \times J}$. This reshaped input is then fed into two spatial HyperGCNs as follows:
\begin{equation}
    \begin{aligned}
        X_{part} &= \sigma\left(\mathrm{Norm}\left(\mathcal{H}_{part} X_{\ell-1} W_{part}\right)\right), \\
        X_{body} &= \sigma\left(\mathrm{Norm}\left(\mathcal{H}_{body} X_{\ell-1} W_{body}\right)\right),
    \end{aligned}
\end{equation}
where $X_{part}$ and $X_{body}$ denote the features obtained from the part and body scales, respectively. $\mathcal{H}_{part}$ and $\mathcal{H}_{body}$ represent the HyperGCN kernels (Eq. \ref{eq-hgcn}) for the part and body scales. $W_{part}$ and $W_{body}$ are learnable weight matrices.
Next, element-wise summation is used to fuse the features from both branches:
\begin{equation}
    X_{spatial} = X_{part} + X_{body} \label{eq-17}
\end{equation} 
In the process of local temporal feature modeling, we construct the temporal adjacency matrices. Following \cite{mehraban2024motionagformer}, we first compute the similarity between individual joints across temporal frames and select the $k$ nearest neighbors as connected nodes in the graph:
\begin{equation}
    A_{tp} = \text{KNN}_{k=2}(X_{spatial} X_{spatial}^{\top})
\end{equation}
where $X_{spatial} \in \mathbb{R}^{B \times T \times CV}$ is the reshaped result of Eq. \ref{eq-17}, and $A_{tp} \in \mathbb{R}^{B \times T \times T}$ is the temporal adjacency matrix.
The local temporal modeling can be expressed as:
\begin{equation}
    \mathrm{GCN}_{tp}(X_{spatial}) = \sigma\left(\mathrm{Norm}\left(\mathcal{G}_{tp} X_{spatial} W_{tp}\right)\right)
\end{equation}
where $\mathcal{G}_{tp}$ is the GCN kernel (Eq. \ref{eq-gcn}), and $W_{tp}$ is a learnable weight matrix.
Finally, the HyperGCN stream can be represented as:
\begin{equation}
    X^{hg}_{\ell-1} = S^{\ell}_{hg}(X_{\ell-1})
\end{equation}


\noindent \textbf{Adaptive Fusion}. 
Similar to MotionBERT \cite{zhu2023motionbert} and MotionAGFormer \cite{mehraban2024motionagformer}, we employ adaptive fusion to aggregate the extracted features from the two streams. The fusion process is defined as:

\begin{equation}
    \begin{aligned}
         X_{\ell} &= \alpha_m \odot X^m_{\ell-1}  + \alpha_{hg} \odot X^{hg}_{\ell-1}, \\
         \alpha_m, \alpha_{hg}  &= \text{softmax}\left(W_{\ell} [X^m_{\ell-1}, X^{hg}_{\ell-1}]\right)
    \end{aligned}
\end{equation}
where $X_{\ell} \in \mathbb{R}^{B \times T \times J \times D}$ represents the output features of the $\ell$-th HGM block, $W_{\ell}$ is a learnable parameter, and $[,]$ denotes the concatenation operation.

\section{Experiments}
We evaluate our proposed HGMamba on two largescale 3D human pose estimation datasets, i.e., Human3.6M \cite{ionescu2013human3} and MPI-INF-3DHP \cite{mehta2017monocular}.
\subsection{Datasets and Evaluation Metrics}
\noindent\textbf{Human3.6M}
 is a widely utilized indoor dataset for 3D human pose estimation, comprising 3.6 million video frames of 11 individuals engaged in 15 distinct daily activities. To maintain a fair evaluation protocol, we adopt the standard practice of training the model using data from subjects 1, 5, 6, 7, and 8, while testing is performed on data from subjects 9 and 11. In line with prior research \cite{tang20233d, zhao2023poseformerv2, mehraban2024motionagformer}, we employ two evaluation protocols. The first protocol, referred to as P1, calculates the Mean Per Joint Position Error (MPJPE) in millimeters between the predicted pose and the ground-truth pose after aligning their root joints (sacrum). The second protocol, known as P2, evaluates Procrustes-MPJPE, where the predicted and ground-truth poses are aligned through a rigid transformation.

\noindent\textbf{MPI-INF-3DHP}
 is a recently popular dataset collected across three distinct settings: green screen, non-green screen, and outdoor environments. Consistent with previous studies \cite{tang20233d, zhao2023poseformerv2, mehraban2024motionagformer}, the evaluation metrics include Mean Per Joint Position Error (MPJPE), Percentage of Correct Keypoints (PCK) within a 150 mm range, and the Area Under the Curve (AUC).

\subsection{Implementation Details}
\noindent\textbf{Model Varints.}
We create three model configurations, detailed in Tab. \ref{tab-1}. 
The selection of each variant depends on specific application needs, like real-time processing or precise estimations. The motion representation dimension of regression head $D' = 512$ for experiments. 
\begin{table}
\centering
\caption{Details of HGMamba model variants. N: Number
of layers. D: Base channels. T: Number of input frames.} \label{tab-1}
\begin{tabular}{llllll} \hline
Method     & N  & D   & T   & Params & MACs  \\ \hline\hline
HGMamba-XS & 12 & 64  & 27  & 2.8M   &  1.14G     \\
HGMamba-S  & 26 & 64  & 81  & 6.1M   &  8.02G     \\
HGMamba-B  & 16 & 128 & 243 & 14.2M  &  64.5G     \\
 \hline
\end{tabular}
\end{table}

\noindent\textbf{Experimental settings.}
Our model is developed using PyTorch \cite{paszke2017automatic} and runs on ubuntu-20.04 system equipped with a RTX A6000 GPU. To enhance performance, horizontal flipping augmentation is applied during both training and testing, as described in \cite{zhao2023poseformerv2,zhu2023motionbert}. The training process utilizes mini-batches of 16 sequences, and network parameters are optimized with the AdamW optimizer over 90 epochs, incorporating a weight decay of 0.01. The initial learning rate is set to 5e-4 and follows an exponential decay schedule with a decay factor of 0.99. For Human3.6M, we employ 2D pose detection results generated by the Stacked Hourglass network \cite{stackedhourglass} and ground truth 2D pose annotations, consistent with \cite{tang20233d, zhao2023poseformerv2, mehraban2024motionagformer}. Similarly, for MPI-INF-3DHP, we use ground truth 2D detections, aligning with the approaches adopted in comparable baseline methods.

\begin{figure*}[th]
	\centering
		\centering
		\includegraphics[width=\linewidth]{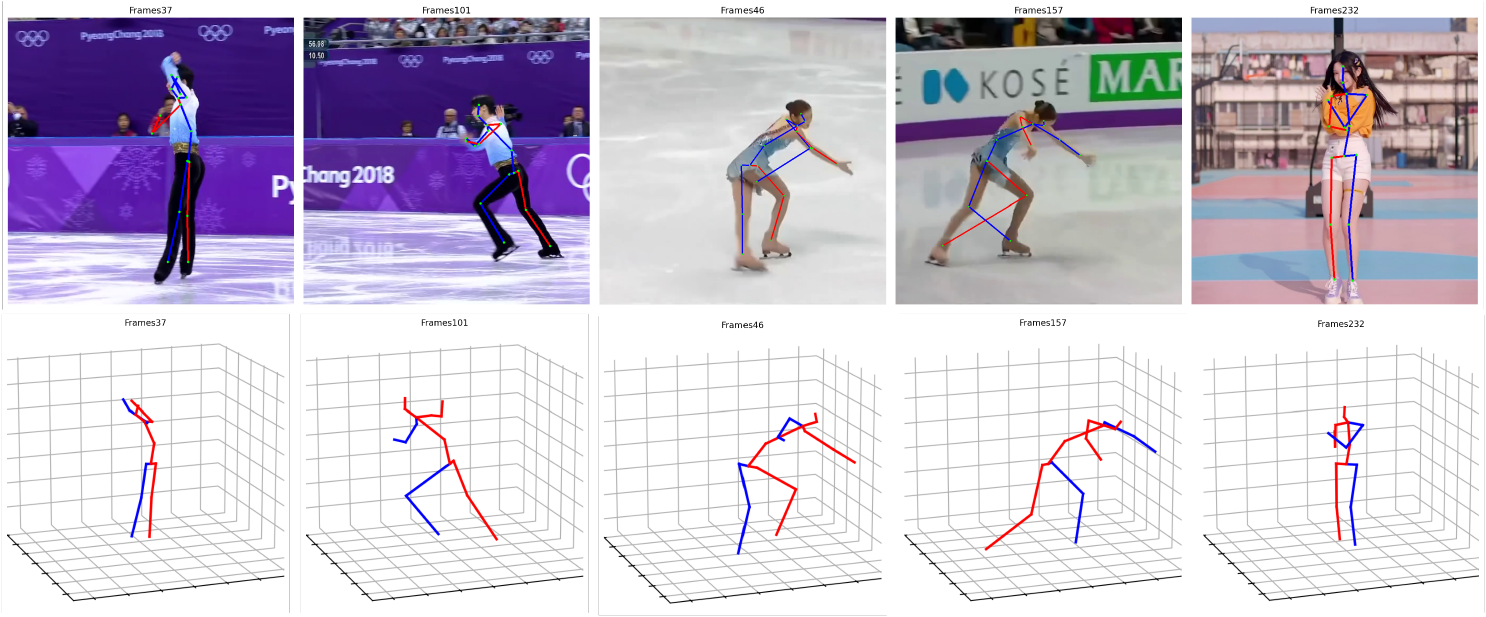}
	\caption{Visualization of 3D pose results from noised 2D Pose.}
	\label{fig:3}
\end{figure*}

\begin{figure*}[t]
	\centering
		\centering
		\includegraphics[width=\linewidth]{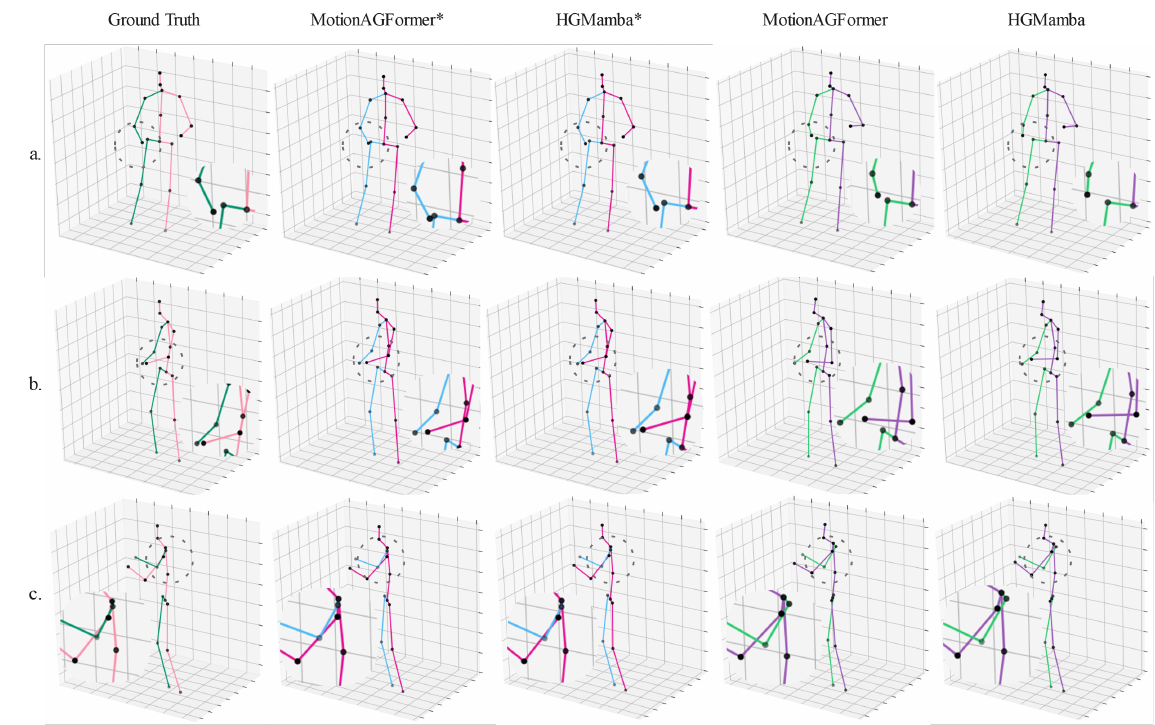}
	\caption{Visualization 3D results for GT-2D and estimated-2D. * denotes result for GT-2D pose. $\dagger$ denotes result for estimated-2D pose by Stacked Hourglass.}
	\label{fig:4}
\end{figure*}

\begin{table*}[h]
\centering
\caption{Quantitative comparisons on Human3.6M. $T$: Number of input frames. $CE$: Estimating center frame only. $P1$: MPJPE
error (mm). $P2$: P-MPJPE error (mm). $P1^\dagger$: P1 error on 2D ground truth. The best and second-best scores are in bold and underlined, respectively.}
\arrayrulecolor{black} \label{tab-2}
\begin{tabular}{lllllllll} \hline
Method                               & Publication & T   & CE       & Param.  & MACs.   & MACs/frame & P1$\downarrow $/P2$\downarrow$ & P1$^\dagger$$\downarrow$  \\ \hline\hline
MHFormer \cite{li2022mhformer}                    & CVPR'22     & 351 & $\checkmark$  & 30.9M   & 7.0G    & 20M        & 43.0/34.4  
    & 30.5                           \\
MixSTE\cite{zhang2022mixste}                               & CVPR'22     & 243 & $\times$ & 33.6M   & 139.G   & 572M       & 40.9/32.6                      & 21.6                           \\
P-STMO\cite{shan2022p}                               & ECCV'22     & 243 & $\checkmark$  & 6.2M    & 0.7G    & 3M         & 42.8/34.4                      & 29.3                           \\
StrideFormer\cite{li2022exploiting}                         & TMM'22      & 351 & $\checkmark$  & 4.0 M   & 0.8 G   & 2 M        & 43.7/35.2                      & 28.5                           \\
Einfalt et al.\cite{einfalt2023uplift}                       & WACV’23     & 351 & $\checkmark$  & 10.4 M  & 0.5 G   & 498 M      & 44.2/35.7                      & -                              \\
STCFormer\cite{tang20233d}                            & CVPR’23     & 243 & $\times$ & 4.7 M  & 19.6 G  & 80 M       & 41.0/\uline{32.0}                      & 21.3                           \\
STCFormer-L\cite{tang20233d}                          & CVPR’23     & 243 & $\times$ & 18.9 M  & 78.2 G  & 321 M      & 40.5/\textbf{31.8}                      & -                              \\
PoseFormerV2\cite{zhao2023poseformerv2}                         & CVPR’23     & 243 & $\checkmark$  & 14.3 M & 0.5 G  & 528 M      & 45.2/35.6                      & -                              \\
GLA-GCN\cite{yu2023gla}                              & ICCV’23     & 243 & $\checkmark$  & 1.3 M  & 1.5 G   & 6 M        & 44.4/34.8                     & 21.0                           \\
MotionBERT\cite{zhu2023motionbert}                           & ICCV’23     & 243 & $\times$ & 42.3 M & 174.8 G & 719M       & 39.2/32.9                     & 17.8                           \\
HDFormer\cite{chen2023hdformer}                             & ICCV’23     & 96  & $\times$ & 3.7 M  & 0.6 G   & 6 M        & 42.6/33.1                      & 21.6                           \\
HSTFormer\cite{qian2023hstformer}                            & arXiv'23    & 81  & $\times$ & 22.7 M & 1.0 G   & 12 M       & 42.7/33.7                      & 27.8                         \\
MotionAGFormer-S\cite{mehraban2024motionagformer}                     & WACV'24     & 81  & $\times$ & 4.8 M   & 6.6 G  & 81 M       & 42.5/35.3                      & 26.5                           \\
MotionAGFormer-L\cite{mehraban2024motionagformer}                     & WACV'24     & 243 & $\times$ & 19.0 M  & 78.3 G  & 322 M     & \textbf{38.4}/32.5         & \uline{17.3}                           \\
HGMamba-XS                           & -           & 27  & $\times$ & 2.8M    & 1.14G        & 42M       & 44.96/38.37                               & 29.51                               \\
HGMamba-S                            & -           & 81  & $\times$ & 6.1M    &  8.02G       & 99M       & 42.84/35.93                    & 22.91                          \\
HGMamba-B                            & -           & 243 & $\times$ & 14.2M   &  64.5G       & 265M       & \uline{38.65}/32.87         & \textbf{13.15}                    \\ \hline
\end{tabular}
\end{table*}

\begin{table}
\centering
\caption{Quantitative comparisons on MPI-INF-3DHP. $T$: Number of input frames. The best and second-best scores are in bold and underlined, respectively.}
\arrayrulecolor{black}\label{tab-3}
\begin{tabular}{lllll} \hline
Method            & T  & PCK$\uparrow $~    & AUC$\uparrow $  & P1(MPJPE)$\downarrow $  \\ \hline\hline
MHFormer\cite{li2022mhformer}          & 9  & 93.8    & 63.3 & 58.0   \\
MixSTE\cite{zhang2022mixste}            & 27 & 94.4    & 66.5 & 54.9   \\
P-STMO\cite{shan2022p}            & 81 & 95.4~ ~ & 75.8 & 32.2   \\
Einfalt et al.\cite{einfalt2023uplift}   & 81 & 95.4    & 67.6 & 46.9   \\
STCFormer\cite{tang20233d}         & 81 & 98.7    & 83.9 & 23.1   \\
PoseFormerV2\cite{zhao2023poseformerv2}      & 81 & 97.9    & 78.8 & 27.8   \\
GLA-GCN\cite{yu2023gla}           & 81 & 98.5    & 79.1 & 27.7   \\
HSTFormer\cite{qian2023hstformer}         & 81 & 97.3    & 71.5 & 41.4   \\
HDFormer\cite{chen2023hdformer}          & 96 & 98.7    & 72.9 & 37.2   \\
MotionBERT\cite{zhu2023motionbert}         & 243 &88.0 &\textbf{99.1} &18.2 \\
MotionAGFormer-XS\cite{mehraban2024motionagformer} & 27 & 98.2    & 83.5 & 19.2   \\
MotionAGFormer-S\cite{mehraban2024motionagformer}  & 81 & 98.3    & 84.5 & 17.1   \\
MotionAGFormer-L\cite{mehraban2024motionagformer}  & 81 & 98.2    & 85.3 & 16.2   \\
HumMUSS\cite{mondal2024hummuss}           & 243 & \textbf{99.0}    & 87.1 & 18.7   \\
HGMamba-XS        & 27 &  98.1    &83.9      & 17.42  \\
HGMamba-S         & 81 &  \uline{98.9}    &86.8      & \uline{15.51}  \\
HGMamba-B         & 81 &  98.7    & \uline{87.9}     & \textbf{14.33}  \\ \hline
\end{tabular}
\end{table}

\subsection{Performance Comparison on Human3.6M}
We conducted a comparative analysis of the HGMamba model against other models using the Human3.6M dataset. To ensure fairness in the evaluation, we did not use any additional data for pre-training. As shown in Table \ref{tab-2}, the results indicate that HGMamba-B achieves a P1 error of 38.65 mm for the estimated 2D pose (using the Stacked Hourglass Network \cite{stackedhourglass}) and 13.15 mm for the ground truth 2D pose. 
Compared to the previous state-of-the-art (SOTA) model, MotionBERT \cite{zhu2023motionbert}, HGMamba-B achieves these results using only 37\% of the computational resources, with an improvement of 0.55 mm for the estimated 2D pose and 4.65 mm for the ground truth 2D pose.
Furthermore, when compared to the latest SOTA model, MotionAGFormer \cite{mehraban2024motionagformer}, HGMamba-B reaches similar results while utilizing only 82\% of the computational resources. Although the P1 error for the estimated 2D pose is slightly higher by 0.25 mm, the P1 error for the ground truth 2D pose is significantly reduced by 4.15 mm.

\subsection{Performance Comparison on MPI-INF-3DHP}
As shown in Table \ref{tab-3}, we evaluated our method on the MPI-INF-3DHP dataset by adjusting the input sequence length of the HGMamba-B model to 81 frames. With a sequence length of 27 frames, the P1 error of HGMamba-XS is 17.42 mm, which is 1.78 mm lower than that of MotionAGFormer-XS. For a sequence length of 81 frames, the P1 errors of HGMamba-S and HGMamba-B are 15.51 mm and 14.33 mm, respectively, outperforming MotionAGFormer-S by 1.95 mm and MotionAGFormer-L by 1.87 mm.

\subsection{Ablation Studies}
To assess the impact of each component in the model and evaluate the performance under different settings, we conducted ablation experiments using the Human3.6M estimated 2D pose dataset (Stacked Hourglass Network).

\noindent\textbf{Effect of Each Component.}
\begin{table}
\centering
\caption{Ablation study for each component used in our method. The evaluation is performed on Human3.6M with MPJPE (mm). The best score is in bold. * indicates mamba stream is attention stream from \cite{mehraban2024motionagformer}. $\dagger$ indicates hyper gcn stream is method \#c.}
\resizebox{0.48\textwidth}{!}{
\arrayrulecolor{black} \label{tab-4}
\begin{tabular}{clllll} \hline
Method     & \multicolumn{4}{c}{Hyper GCN stream*~}                    & \multirow{2}{*}{P1(MPJPE)$\downarrow $}  \\ \cline{1-5}
Components & GCN+TCN      & B-HyperGCN   & P-HyperGCN   & T-GCN        &                             \\ \hline\hline
baseline   & $\checkmark$ &              &              &              &   39.11                          \\
\#a          &              & $\checkmark$ &              & $\checkmark$ & 38.96                            \\
\#b          &              &              & $\checkmark$ & $\checkmark$ &  39.21                           \\
\#c          &              & $\checkmark$ & $\checkmark$ & $\checkmark$ &  \textbf{38.74}                           \\ \hline\hline
Method     & \multicolumn{4}{c}{Mamba stream$^\dagger$}                         & \multirow{2}{*}{P1(MPJPE)$\downarrow $}  \\ \cline{1-5}
Components & ST-Mamba     & TS-Mamba     & Forward      & Backward     &                             \\ \hline\hline
\#d          & $\checkmark$ &              & $\checkmark$ &              &   39.46                          \\
\#e          & $\checkmark$ &              & $\checkmark$ & $\checkmark$ &   38.88                          \\
\#f          &              & $\checkmark$ & $\checkmark$ & $\checkmark$ &   38.92                          \\
\#g1         & $\checkmark$ & $\checkmark$ & $\checkmark$ & $\checkmark$ &   38.71                          \\
\#g2         & $\checkmark$ & $\checkmark$ & $\checkmark$ & $\checkmark$ &  \textbf{38.65}                           \\ \hline
\end{tabular}
\arrayrulecolor{black}
}
\end{table}
As shown in Table \ref{tab-4}, we use the GCN and Attention stream architectures from \cite{mehraban2024motionagformer} as a baseline. First, we replace the GCN stream with body-scale Hyper-GCN and part-scale Hyper-GCN. Our experiments show no significant improvement when either Hyper-GCN is used independently. However, when both are combined, the P1 error decreases from 39.11 mm to 38.74 mm. This demonstrates the importance of incorporating multi-granularity body part semantics in local feature modeling.

Next, we keep the Hyper-GCN stream unchanged and replace the attention stream in the baseline with a Mamba stream. We observe that when using either the spatial-first Mamba (\# e) or temporal-first Mamba (\# f) individually, performance is degraded compared to the original attention stream. However, when combining the two in series (\# g1, ST-Mamba $\rightarrow$ TS-Mamba), the P1 error improves slightly from 38.74 mm to 38.71 mm. When combining them in parallel (\# g2), the P1 error further decreases to 38.65 mm.

\noindent\textbf{Parameter Setting Analysis.}
%
\begin{table}
\centering
\caption{Ablation study for hyper-parameter setting in layer depth (N), dimension (D) and frame length (T). The evaluation is performed on Human3.6M with MPJPE (mm). The best score is in bold. }  \label{tab-5}
\arrayrulecolor{black}
\begin{tabular}{llll} \hline
\multicolumn{1}{!{\color{black}}l}{Depth (N)~} & \multicolumn{1}{l!{\color{black}}}{Dimension (D)~} & Input Length (T) & P1(MPJPE)$\downarrow $  \\ \hline\hline
4                                                    & 64                                                       & 27               &   46.17         \\
8                                                    & 64                                                       & 27               &   45.85         \\
12                                                   & 64                                                       & 27               &   44.96         \\ \hline
16                                                   & 64                                                       & 81               &   43.57         \\
20                                                   & 64                                                       & 81               &   43.12         \\
26                                                   & 64                                                       & 81               &   42.84        \\ \hline
4                                                    & 128                                                      & 243              &   39.54         \\
8                                                    & 128                                                      & 243              &   38.78         \\
16                                                   & 128                                                      & 243              &   \textbf{38.65}         \\ \hline
\end{tabular}
\arrayrulecolor{black}
\end{table}
%
%
Tab. \ref{tab-5} shows the effect of different hyperparameter settings on the P1 error in our method. The three main hyperparameters are the depth of the network (N), the dimensions of the underlying channels (D), and the length of the input 2D Pose sequence (T).
We set the configurations into 3 groups, the first group keeps D=64, T=27, to find the optimal depth configuration for short video sequences. The second group keeps D=64, T=81 to find the optimal depth configuration for medium video sequences. The third group maintains D=128, T=243 to seek the optimal depth configuration under long video sequences. Finally we got the three variants in Tab. \ref{tab-1}, depending on the parameters and performance.


\subsection{Qualitative Analysis}


\noindent\textbf{Result Visualization with Noisy 2D Poses.} 
As shown in Fig. \ref{fig:3}, we evaluate the robustness of our model to noisy 2D poses. We use HGMamba-B to perform 3D pose estimation based on 2D poses detected by HRNet \cite{sun2019deep} on various objects. For this evaluation, we intentionally select 2D pose samples with obvious errors. The results demonstrate that, even when 2D poses are inaccurately estimated, our model can still accurately predict the correct 3D pose.


\noindent\textbf{Result Visalization with GT-2D and Estimated-2D.}
As shown in Tab. \ref{tab-2}, our model achieves a more significant reduction in P1 error for ground truth 2D poses (GT-2D) compared to estimated 2D poses (estimated-2D).

To further illustrate the performance difference between GT-2D and estimated-2D, we randomly select three samples from the Human3.6M evaluation set, each with two inputs (Ground Truth 2D and Estimated-2D). We then predict the corresponding 3D poses using four models: MotionAGFormer-L for both GT-2D and estimated-2D, and HGMamba-B for both GT-2D and estimated-2D. 

From the results in Fig. \ref{fig:4}, we observe that our model yields similar results to MotionAGFormer for estimated-2D poses, but provides more accurate predictions for GT-2D poses. This suggests that a more accurate 2D pose detector can significantly enhance the performance of our model.


\section{Conclusion}
We present HGMamba, an approach that leverages Hyper-GCN to capture local joint structural information and combines it with the latest state-of-the-art model, Mamba, which effectively captures global joint interdependencies. This fusion enhances the model's ability to understand the 3D structure inherent in input 2D sequences. Additionally, HGMamba offers multiple adaptive variants, allowing for the optimal balance between speed and accuracy. Empirical evaluations demonstrate that our approach achieves state-of-the-art results on both the Human3.6M and MPI-INF-3DHP datasets with ground truth 2D pose data, while maintaining competitive performance on estimated 2D poses.

\bibliographystyle{IEEEtran}
\bibliography{mybibfile}

\begin{thebibliography}{10}
\providecommand{\url}[1]{#1}
\csname url@samestyle\endcsname
\providecommand{\newblock}{\relax}
\providecommand{\bibinfo}[2]{#2}
\providecommand{\BIBentrySTDinterwordspacing}{\spaceskip=0pt\relax}
\providecommand{\BIBentryALTinterwordstretchfactor}{4}
\providecommand{\BIBentryALTinterwordspacing}{\spaceskip=\fontdimen2\font plus
\BIBentryALTinterwordstretchfactor\fontdimen3\font minus \fontdimen4\font\relax}
\providecommand{\BIBforeignlanguage}[2]{{%
\expandafter\ifx\csname l@#1\endcsname\relax
\typeout{** WARNING: IEEEtran.bst: No hyphenation pattern has been}%
\typeout{** loaded for the language `#1'. Using the pattern for}%
\typeout{** the default language instead.}%
\else
\language=\csname l@#1\endcsname
\fi
#2}}
\providecommand{\BIBdecl}{\relax}
\BIBdecl

\bibitem{baltieri20113dpes}
D.~Baltieri, R.~Vezzani, and R.~Cucchiara, ``3dpes: 3d people dataset for surveillance and forensics,'' in \emph{Proceedings of the 2011 joint ACM workshop on Human gesture and behavior understanding}, 2011, pp. 59--64.

\bibitem{cui2024stsd}
H.~Cui and T.~Hayama, ``Stsd: spatial--temporal semantic decomposition transformer for skeleton-based action recognition,'' \emph{Multimedia Systems}, vol.~30, no.~1, p.~43, 2024.

\bibitem{chung2022comparative}
J.-L. Chung, L.-Y. Ong, and M.-C. Leow, ``Comparative analysis of skeleton-based human pose estimation,'' \emph{Future Internet}, vol.~14, no.~12, p. 380, 2022.

\bibitem{chen2021channel}
Y.~Chen, Z.~Zhang, C.~Yuan, B.~Li, Y.~Deng, and W.~Hu, ``Channel-wise topology refinement graph convolution for skeleton-based action recognition,'' in \emph{Proceedings of the IEEE/CVF international conference on computer vision}, 2021, pp. 13\,359--13\,368.

\bibitem{cui2022lgst}
H.~Cui, R.~Huang, R.~Zhang, and C.~Huang, ``Lgst-drop: label-guided structural dropout for spatial--temporal convolutional neural networks,'' \emph{Journal of Electronic Imaging}, vol.~31, no.~3, pp. 033\,036--033\,036, 2022.

\bibitem{yan2018spatial}
S.~Yan, Y.~Xiong, and D.~Lin, ``Spatial temporal graph convolutional networks for skeleton-based action recognition,'' in \emph{Proceedings of the AAAI conference on artificial intelligence}, vol.~32, no.~1, 2018.

\bibitem{cui2024joint}
H.~Cui and T.~Hayama, ``Joint-partition group attention for skeleton-based action recognition,'' \emph{Signal Processing}, vol. 224, p. 109592, 2024.

\bibitem{sarafianos20163d}
N.~Sarafianos, B.~Boteanu, B.~Ionescu, and I.~A. Kakadiaris, ``3d human pose estimation: A review of the literature and analysis of covariates,'' \emph{Computer Vision and Image Understanding}, vol. 152, pp. 1--20, 2016.

\bibitem{he2017mask}
K.~He, G.~Gkioxari, P.~Doll{\'a}r, and R.~Girshick, ``Mask r-cnn,'' in \emph{Proceedings of the IEEE international conference on computer vision}, 2017, pp. 2961--2969.

\bibitem{chen2018cascaded}
Y.~Chen, Z.~Wang, Y.~Peng, Z.~Zhang, G.~Yu, and J.~Sun, ``Cascaded pyramid network for multi-person pose estimation,'' in \emph{Proceedings of the IEEE conference on computer vision and pattern recognition}, 2018, pp. 7103--7112.

\bibitem{stackedhourglass}
A.~Newell, K.~Yang, and J.~Deng, ``Stacked hourglass networks for human pose estimation,'' in \emph{European Conference on Computer Vision}.\hskip 1em plus 0.5em minus 0.4em\relax Springer, 2016, pp. 483--499.

\bibitem{sun2019deep}
K.~Sun, B.~Xiao, D.~Liu, and J.~Wang, ``Deep high-resolution representation learning for human pose estimation,'' in \emph{Proceedings of the IEEE/CVF conference on computer vision and pattern recognition}, 2019, pp. 5693--5703.

\bibitem{maji2022yolo}
D.~Maji, S.~Nagori, M.~Mathew, and D.~Poddar, ``Yolo-pose: Enhancing yolo for multi person pose estimation using object keypoint similarity loss,'' in \emph{Proceedings of the IEEE/CVF Conference on Computer Vision and Pattern Recognition}, 2022, pp. 2637--2646.

\bibitem{lugaresi2019mediapipe}
C.~Lugaresi, J.~Tang, H.~Nash, C.~McClanahan, E.~Uboweja, M.~Hays, F.~Zhang, C.-L. Chang, M.~Yong, J.~Lee \emph{et~al.}, ``Mediapipe: A framework for perceiving and processing reality,'' in \emph{Third workshop on computer vision for AR/VR at IEEE computer vision and pattern recognition (CVPR)}, vol. 2019, 2019.

\bibitem{zhang2022mixste}
J.~Zhang, Z.~Tu, J.~Yang, Y.~Chen, and J.~Yuan, ``Mixste: Seq2seq mixed spatio-temporal encoder for 3d human pose estimation in video,'' in \emph{Proceedings of the IEEE/CVF conference on computer vision and pattern recognition}, 2022, pp. 13\,232--13\,242.

\bibitem{yu2023gla}
B.~X. Yu, Z.~Zhang, Y.~Liu, S.-h. Zhong, Y.~Liu, and C.~W. Chen, ``Gla-gcn: Global-local adaptive graph convolutional network for 3d human pose estimation from monocular video,'' in \emph{Proceedings of the IEEE/CVF International Conference on Computer Vision}, 2023, pp. 8818--8829.

\bibitem{li2022mhformer}
W.~Li, H.~Liu, H.~Tang, P.~Wang, and L.~Van~Gool, ``Mhformer: Multi-hypothesis transformer for 3d human pose estimation,'' in \emph{Proceedings of the IEEE/CVF Conference on Computer Vision and Pattern Recognition}, 2022, pp. 13\,147--13\,156.

\bibitem{einfalt2023uplift}
M.~Einfalt, K.~Ludwig, and R.~Lienhart, ``Uplift and upsample: Efficient 3d human pose estimation with uplifting transformers,'' in \emph{Proceedings of the IEEE/CVF Winter Conference on Applications of Computer Vision}, 2023, pp. 2903--2913.

\bibitem{tang20233d}
Z.~Tang, Z.~Qiu, Y.~Hao, R.~Hong, and T.~Yao, ``3d human pose estimation with spatio-temporal criss-cross attention,'' in \emph{Proceedings of the IEEE/CVF Conference on Computer Vision and Pattern Recognition}, 2023, pp. 4790--4799.

\bibitem{zhao2023poseformerv2}
Q.~Zhao, C.~Zheng, M.~Liu, P.~Wang, and C.~Chen, ``Poseformerv2: Exploring frequency domain for efficient and robust 3d human pose estimation,'' in \emph{Proceedings of the IEEE/CVF Conference on Computer Vision and Pattern Recognition}, 2023, pp. 8877--8886.

\bibitem{qian2023hstformer}
X.~Qian, Y.~Tang, N.~Zhang, M.~Han, J.~Xiao, M.-C. Huang, and R.-S. Lin, ``Hstformer: Hierarchical spatial-temporal transformers for 3d human pose estimation,'' \emph{arXiv preprint arXiv:2301.07322}, 2023.

\bibitem{chen2023hdformer}
H.~Chen, J.-Y. He, W.~Xiang, Z.-Q. Cheng, W.~Liu, H.~Liu, B.~Luo, Y.~Geng, and X.~Xie, ``Hdformer: high-order directed transformer for 3d human pose estimation,'' in \emph{Proceedings of the Thirty-Second International Joint Conference on Artificial Intelligence}, 2023, pp. 581--589.

\bibitem{zhu2023motionbert}
W.~Zhu, X.~Ma, Z.~Liu, L.~Liu, W.~Wu, and Y.~Wang, ``Motionbert: A unified perspective on learning human motion representations,'' in \emph{Proceedings of the IEEE/CVF International Conference on Computer Vision}, 2023, pp. 15\,085--15\,099.

\bibitem{li2022exploiting}
W.~Li, H.~Liu, R.~Ding, M.~Liu, P.~Wang, and W.~Yang, ``Exploiting temporal contexts with strided transformer for 3d human pose estimation,'' \emph{IEEE Transactions on Multimedia}, vol.~25, pp. 1282--1293, 2022.

\bibitem{mehraban2024motionagformer}
S.~Mehraban, V.~Adeli, and B.~Taati, ``Motionagformer: Enhancing 3d human pose estimation with a transformer-gcnformer network,'' in \emph{Proceedings of the IEEE/CVF Winter Conference on Applications of Computer Vision}, 2024, pp. 6920--6930.

\bibitem{pavllo20193d}
D.~Pavllo, C.~Feichtenhofer, D.~Grangier, and M.~Auli, ``3d human pose estimation in video with temporal convolutions and semi-supervised training,'' in \emph{Proceedings of the IEEE/CVF conference on computer vision and pattern recognition}, 2019, pp. 7753--7762.

\bibitem{liu2020attention}
R.~Liu, J.~Shen, H.~Wang, C.~Chen, S.-c. Cheung, and V.~Asari, ``Attention mechanism exploits temporal contexts: Real-time 3d human pose reconstruction,'' in \emph{Proceedings of the IEEE/CVF conference on computer vision and pattern recognition}, 2020, pp. 5064--5073.

\bibitem{hu2021conditional}
W.~Hu, C.~Zhang, F.~Zhan, L.~Zhang, and T.-T. Wong, ``Conditional directed graph convolution for 3d human pose estimation,'' in \emph{Proceedings of the 29th ACM International Conference on Multimedia}, 2021, pp. 602--611.

\bibitem{bai2021hypergraph}
S.~Bai, F.~Zhang, and P.~H. Torr, ``Hypergraph convolution and hypergraph attention,'' \emph{Pattern Recognition}, vol. 110, p. 107637, 2021.

\bibitem{zhu2022selective}
Y.~Zhu, G.~Huang, X.~Xu, Y.~Ji, and F.~Shen, ``Selective hypergraph convolutional networks for skeleton-based action recognition,'' in \emph{Proceedings of the 2022 international conference on multimedia retrieval}, 2022, pp. 518--526.

\bibitem{mamba}
A.~Gu and T.~Dao, ``Mamba: Linear-time sequence modeling with selective state spaces,'' \emph{arXiv preprint arXiv:2312.00752}, 2023.

\bibitem{mamba2}
T.~Dao and A.~Gu, ``Transformers are {SSM}s: Generalized models and efficient algorithms through structured state space duality,'' in \emph{International Conference on Machine Learning (ICML)}, 2024.

\bibitem{zhu2024vision}
L.~Zhu, B.~Liao, Q.~Zhang, X.~Wang, W.~Liu, and X.~Wang, ``Vision mamba: Efficient visual representation learning with bidirectional state space model,'' \emph{arXiv preprint arXiv:2401.09417}, 2024.

\bibitem{pavlakos2018ordinal}
G.~Pavlakos, X.~Zhou, and K.~Daniilidis, ``Ordinal depth supervision for 3d human pose estimation,'' in \emph{Proceedings of the IEEE conference on computer vision and pattern recognition}, 2018, pp. 7307--7316.

\bibitem{mondal2024hummuss}
A.~Mondal, S.~Alletto, and D.~Tome, ``Hummuss: Human motion understanding using state space models,'' in \emph{Proceedings of the IEEE/CVF Conference on Computer Vision and Pattern Recognition}, 2024, pp. 2318--2330.

\bibitem{gu2022parameterization}
A.~Gu, K.~Goel, A.~Gupta, and C.~R{\'e}, ``On the parameterization and initialization of diagonal state space models,'' \emph{Advances in Neural Information Processing Systems}, vol.~35, pp. 35\,971--35\,983, 2022.

\bibitem{gu2021efficiently}
A.~Gu, K.~Goel, and C.~R{\'e}, ``Efficiently modeling long sequences with structured state spaces,'' \emph{arXiv preprint arXiv:2111.00396}, 2021.

\bibitem{gupta2022diagonal}
A.~Gupta, A.~Gu, and J.~Berant, ``Diagonal state spaces are as effective as structured state spaces,'' \emph{Advances in Neural Information Processing Systems}, vol.~35, pp. 22\,982--22\,994, 2022.

\bibitem{mehta2022long}
H.~Mehta, A.~Gupta, A.~Cutkosky, and B.~Neyshabur, ``Long range language modeling via gated state spaces,'' \emph{arXiv preprint arXiv:2206.13947}, 2022.

\bibitem{zhang2018shufflenet}
X.~Zhang, X.~Zhou, M.~Lin, and J.~Sun, ``Shufflenet: An extremely efficient convolutional neural network for mobile devices,'' in \emph{Proceedings of the IEEE conference on computer vision and pattern recognition}, 2018, pp. 6848--6856.

\bibitem{huang2021shuffle}
Z.~Huang, Y.~Ben, G.~Luo, P.~Cheng, G.~Yu, and B.~Fu, ``Shuffle transformer: Rethinking spatial shuffle for vision transformer,'' \emph{arXiv preprint arXiv:2106.03650}, 2021.

\bibitem{huang2024stochastic}
Z.~Huang, H.~Chen, J.~Li, J.~Lan, H.~Zhu, W.~Wang, and L.~Wang, ``Stochastic layer-wise shuffle: A good practice to improve vision mamba training,'' \emph{arXiv preprint arXiv:2408.17081}, 2024.

\bibitem{feng2019hypergraph}
Y.~Feng, H.~You, Z.~Zhang, R.~Ji, and Y.~Gao, ``Hypergraph neural networks,'' in \emph{Proceedings of the AAAI conference on artificial intelligence}, vol.~33, no.~01, 2019, pp. 3558--3565.

\bibitem{jiang2019dynamic}
J.~Jiang, Y.~Wei, Y.~Feng, J.~Cao, and Y.~Gao, ``Dynamic hypergraph neural networks.'' in \emph{IJCAI}, 2019, pp. 2635--2641.

\bibitem{hendrycks2016gaussian}
D.~Hendrycks and K.~Gimpel, ``Gaussian error linear units (gelus),'' \emph{arXiv preprint arXiv:1606.08415}, 2016.

\bibitem{ionescu2013human3}
C.~Ionescu, D.~Papava, V.~Olaru, and C.~Sminchisescu, ``Human3. 6m: Large scale datasets and predictive methods for 3d human sensing in natural environments,'' \emph{IEEE transactions on pattern analysis and machine intelligence}, vol.~36, no.~7, pp. 1325--1339, 2013.

\bibitem{mehta2017monocular}
D.~Mehta, H.~Rhodin, D.~Casas, P.~Fua, O.~Sotnychenko, W.~Xu, and C.~Theobalt, ``Monocular 3d human pose estimation in the wild using improved cnn supervision,'' in \emph{2017 international conference on 3D vision (3DV)}.\hskip 1em plus 0.5em minus 0.4em\relax IEEE, 2017, pp. 506--516.

\bibitem{paszke2017automatic}
A.~Paszke, S.~Gross, S.~Chintala, G.~Chanan, E.~Yang, Z.~DeVito, Z.~Lin, A.~Desmaison, L.~Antiga, and A.~Lerer, ``Automatic differentiation in pytorch,'' 2017.

\bibitem{shan2022p}
W.~Shan, Z.~Liu, X.~Zhang, S.~Wang, S.~Ma, and W.~Gao, ``P-stmo: Pre-trained spatial temporal many-to-one model for 3d human pose estimation,'' in \emph{European Conference on Computer Vision}.\hskip 1em plus 0.5em minus 0.4em\relax Springer, 2022, pp. 461--478.

\end{thebibliography}
\end{document}